# AN AGENT BASED ARCHITECTURE (USING PLANNING) FOR DYNAMIC AND SEMANTIC WEB SERVICES COMPOSITION IN AN ebXML CONTEXT

O.Hioual[1] and Z.Boufaida[2]

LIRE Laboratory, Mentouri University of
Constantine, Ain ElBey 25000, Algeria
{¹ouassila.hioual,²zboufaida}@gmail.com

## ABSTRACT

*The process-based semantic composition of Web Services is gaining a considerable momentum as an approach for the effective integration of distributed, heterogeneous, and autonomous applications. To compose Web Services semantically, we need an ontology. There are several ways of inserting semantics in Web Services. One of them consists of using description languages like OWL-S. In this paper, we introduce our work which consists in the proposition of a new model and the use of semantic matching technology for semantic and dynamic composition of ebXML business processes.*

## 1. INTRODUCTION

The great success of Web services, due especially to their richness of application made possible by open common standards, has led to their wide proliferation and a tremendous variety of Web services (WS) are now available. However, this proliferation has rendered the discovery, search and use of an appropriate Web services arduous. These tasks are increasingly complicated, especially while dealing with composite Web service to response to an ostensible long-term complex user's goal. The automatic Web service composition task consists of finding an appropriate combination of existing Web services to achieve a global goal. Solving this problem involves mixing and matching component Web services according to certain features. These features can be divided into two main groups [1]:
- Features related to the user, including the user's constraints and preferences.
- Features related to Web services and which can be divided into two subgroups, *internal* and *external* features. *Internal* features include quality of service (QoS) attributes, and *external* features include existing restrictions on the connection of Web services, (e.g., a hotel room should be reserved for a conference usually after booking the flight). *External* features are specified in the Web service ontology language, OWL-S [2], through a set of control constructs such as, *Sequence*, *Unordered*, *Choice*, etc.

In the field of Web services, the Semantic Web Services (SWS) approach [3] is a step toward dynamic service discovery and composition [4, 5] where intelligent systems try to build service compositions from abstract user requirements without a manual selection of services. SWS build on knowledge representation techniques, with ontologies describing a domain in a formal manner, and AI planning methods to make composition systems more autonomous.

An agent possesses the ability to understand and interact with its environment. Because of being context- aware, autonomous and able to interpret semantics with the help of ontological knowledge representation, agents are a necessary complement to **web services** to realize the vision of semantic web [6]. The relation between WS and agent systems has already been mentioned by [7] where a web service is viewed as ''an abstract notion that must be implemented by an agent". Several arguments have been made to support the idea of integration of WS and agent infrastructure, including [8, 9, 10] but perhaps none more evocative than statements made in [11] which clearly expresses the notion that, "software agents are the running programs that drive WS - both to implement them and to access them as computational resources that act on behalf of a person or organization". To enable this integration, several core issues are there out of which bidirectional service discovery, service invocation and composition are the most pertinent. Some significant work has begun taking place in the research community as regards composition of web services from agent infrastructure. Our goal is to take the flexible interaction schemes from the *Multi- Agent Systems* (MAS) research, and utilize them to enable composition among SWS, a paradigm that supports rigid and mechanical interaction protocols, and agent infrastructure. In this paper, we propose an agent based architecture for conducting such composition.

The combination of Web services has attracted the interest of many researchers [12, 13, 14, 15]. The literature [16, 17, 18] demonstrates that automatic planning is an interesting tool for dynamic and automatic web services composition. However, the most proposed architectures repose on a centralized composition.
Our main goal is to provide an agent based architecture that allows composing web services, stocked in an ebXML registry, dynamically. The proposed composition mode is based on multi-agent planning [19, 20].

According to [20], Web service characteristics are very close to those of an agent within the framework of the multi-agent planning: it's autonomous and can communicate with the other Web services. Furthermore, it is possible, by adding to it a semantic description of its features, to argue about its capacities.

Two distributed planning approaches exist. In the first one, each agent produces a plan. The goal is to coordinate these different plans in order to avoid conflicts: it's called the plans coordination. In the second approach, the objective is to allow agents to co-build a plan by taking into account each agent's competences as one goes along co-construction: it's the dialectic plans synthesis. In this approach, agents have the possibility of sending hypothesizes in order to not block the dialectic process when they are in a blocked situation.

In this paper, we present a semantic web services composition architecture based on MAS. The semantic for service description is based on sub-ontologies and the responsibility of composing and coordinating the execution of a composite service specified by a user is centralized around one component called composer agent. Another agent called the general manager agent is responsible for global control of agents' tasks. The composer agent is responsible for global control of composition tasks in conjunction with other components called manager agents; it implements many strategies of composition control constructions (sequence, split, If-Then-Else, etc.). The proposed web services composition method is based on the dialectic plans synthesis. Manager agents (Web services) propose there competences to the composer agent in order to achieve the fixed user objectives by sending, if necessary, hypothesizes on incomplete data.

The rest of the paper is organized as follows. In section 2, we give an overview of existing researches. Section 3 highlights the merger of different candidate technologies (Web services and Multi-Agent System). Section 4 briefly defines the problem and the objectives aimed by our research. We introduce the general agent based architecture then we detail its functioning and architecture of every agent that composes it in section 5. Section 6 highlights conclusion and intended future work followed by references.

## 2. RELATED WORK

Various composition techniques of Web Services exist in the literature. There are two groups : the static techniques i.e., which are defined by means of Business Process (orchestration and choreography); and the dynamic ones, in which the composition of Web services takes into account available services, their features and purpose to be reached.

Techniques of dynamic composition can be grouped in two sub-families: techniques using an approach based on workflows (BPs) and those based on techniques on artificial intelligence. In our work, we are interested in the second type of composition in which preferences and constraints of the customers will be considered.

Several works have proposed different automated planning techniques to address the problem of automated composition (see, e.g., [12, 21, 22, 5]). In this paper, we are interested to this type of research works. These works lean essentially on the classic planning [22, 23, 24], the planning was based on rules [25] and the hierarchical planning [5]. Most of these works advanced the fact that the syntactic description (WSDL) of the services is not sufficient and proposed solutions based on a semantic description (OWL-S).

The collective planning, in a multi-agents context, is also a promising solution because of the correspondence between the notion of atomic and composite process of OWL-S and that of operator and of method in the planning. The objective is to have agents capable of collaborating to realize a common purpose.

In distributed planning, the domain of planning is distributed on all agents. Every agent is capable of realizing certain number of actions: its competences. It is the pooling and the organization of the competences of every agent, with the aim of resolving the given problem that is going to allow bringing to the foreground a solution plan. This type of planning is the one which interests us most, to build the model of composition of the Web services in the architecture on agents' proposed base. The peculiarity of our composition model lives in the fact that the outcome to a solution plan is not totally distributed because we use an agent composer allowing to collaborate with the various manager agents (reserved Web services). The idea is that there is always an agent capable of overseeing the state of progress of the process of composition and of being able to localize the problem in case of failure.

## 3. USING A MULTI-AGENT SYSTEM AS A DYNAMIC SERVICE COMPOSITION INFRASTRUCTURE

A Multi-Agent System (MAS) is a distributed system composed of autonomous entities, called agents. These agents need to interact and cooperate in order to achieve global tasks. One of the main properties of MAS is that it relies on the distribution of cooperation algorithms rather than on centralized processes. We underline two main features of MAS with regard to dynamic service composition.

MAS enable complex interactions between entities, using high level semantic languages. This feature seems essential in environments dealing with *various, heterogeneous information* from physical sensors, services or users preferences. Integration of such data is only possible at a higher level where all kind of information (about services, context ...) is expressed semantically.

In MAS, autonomous entities with limited capabilities coordinate in order to achieve complex tasks. *Emergent coordination and flexible organization* patterns enable groups of agents to create and reconfigure application dynamically depending on conditions. Such patterns seem well adapted to dynamic composition of elementary functionalities in an open, dynamic environment.

## 4. RESEARCH PROBLEM AND OBJECTIVES

The composition of Web services aims at producing a description specifying a sequence of calls to services as well as the way these services are connected between them, with the aim of resolving a given objective. This operation takes place in three steps:
1- The Web services are looked for and selected from an UDDI directory (in our case from the ebXML registry) according to needs to realize;
2- The composition is made by using the semantic description of the selected services;
3- A description of the composite service, i.e., the sequence of movements of the calls to the selected services, is created.

Our research problem is a part of the semantic and dynamic composition of Web services in the context of ebXML (electronic business eXtensible Markup Language). The goal is to add a component to the functional specification of ebXML, whose role is to:
- Look, semantically, for BPs (Business Processes) that meet the requirements of the client company.
- Combine, semantically, BPs to meet the needs of the client company.

To reach this objective, it is necessary to define an ebXML domain ontology, to propose a semantic and dynamic composition model of the Web services, and to exploit the techniques of semantic matching. In this paper, we are mainly interested in the Web services composition model.

### 4.1. WHERE WE ARE LOCATING THE NEW COMPONENT

ebXML [26] is a set of specifications that together enable a modular electronic business framework. The vision of ebXML is to enable a global electronic marketplace where enterprises of any size and in any geographical location can meet and conduct business with each other through the exchange of XML-based messages.

Figure 1 is an illustration based on the ebXML Technical Architecture Specification [26] which gives an outline of what ebXML means for business.

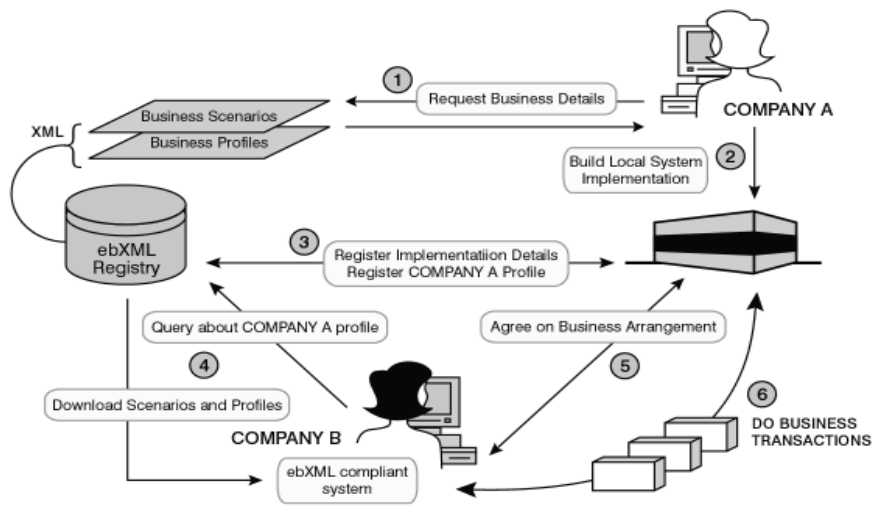

Figure 1. High-level overview of ebXML interaction between two companies [26]

In a context of ebXML, all the Business Processes (BPs) of companies (big or small), using ebXML as tool to participate in global markets, are grouped in a Registry/Repository.

According to our point of view, the problem of ebXML BPs' composition appears in two scenarios:

1. When the process looking for with specified parameters does not exist during the scenario of collaboration between two commercial partners. Combining one ore more services with similar processes can satisfy the needs of the applicant company.

2. When the phase of the parameters negotiation of a business process fails. In order to not stop a possible partnership, a component added to the functional specification of ebXML will look for similar BPs and which compose them. This will be based on the history of the negotiation phase.

The Web services composition problem in an ebXML context appears: during the first phase of the collaboration scenario between two commercial partners. When a company looking for a potential partner consults the ebXML registry, the service asked with certain parameters cannot exist but at the same time, it can exist services, which combined together, can answer its needs.

To resolve this problem, we propose an agent based architecture, allowing composing Web services, according to the functional specification of ebXML. This architecture defines several levels of responsibility and uses the classic planning with the dialectic plans synthesis (planning under hypothesis). The proposed Web services composition model allows passing from the initial state to the final state in order to produce the solution plan, according to the planning domain.

## 5. PROPOSED ARCHITECTURE

As shown it figure 2, the architecture which we propose is an agent based one which defines several levels of responsibility. Four agents' types compose it: the *request re-constructor agent*, the *general manager agent*, the *composer agent* and a set of *manager agents*.

Another important component allows specifying semantics of the various Web services existing at the ebXML registry. It is about a global OWL-S ontology, for the semantic annotation of these services. OWL-S is very rich and imposes only very few constraints on the way of expressing this semantics. Because our work is inspired from that of [19] and [20], we made certain number of hypotheses and limitations on this last one to facilitate the passage of a semantic representation towards a plan:
- The simple process (SimpleProcess) is not treated. Indeed, they correspond only to abstractions of atomic or composite processes;
- OWL-S allows to take into account the indeterminism due to the execution of a service, that is he leaves the possibility of defining the result of a Web service according to his behavior (success, error or absence of answer) via the use of the classes *ConditionalEffect* and *ConditionalOutput* of OWL-S which allow to define the condition under which a result is produced. As the composition takes place before the execution of the service, we authorize the definition only of a single possible result: the result corresponding to the success of the execution of the service;
- We specify the preconditions and the effects of the processes STRIPS in the formalism, see example bellow, to allow their use directly during the creation of the domain;

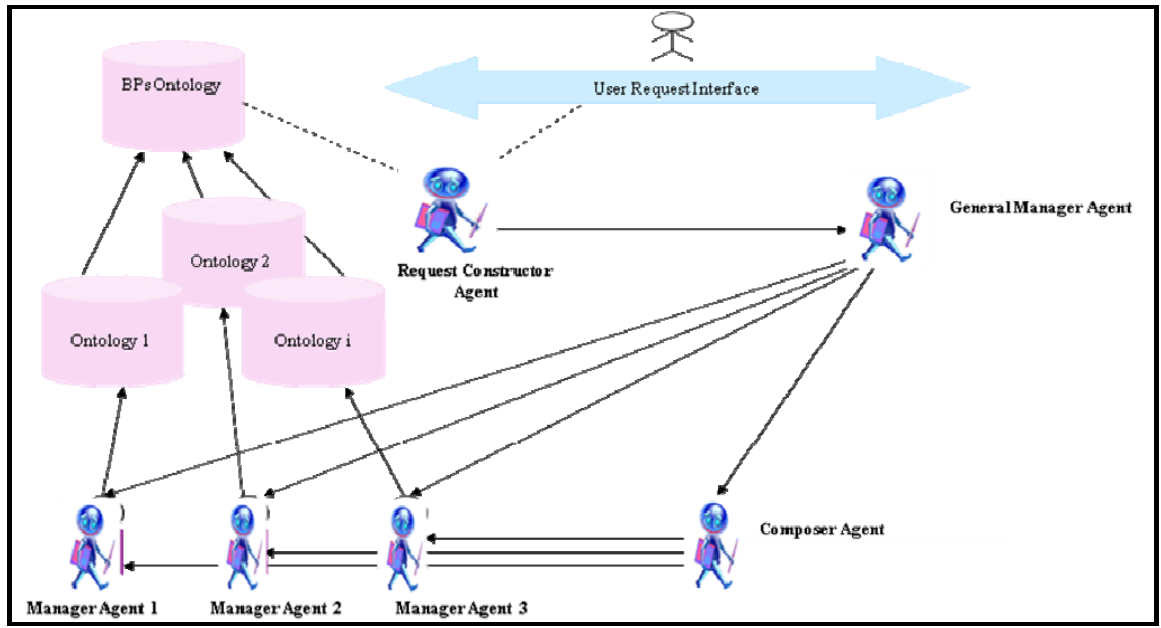

Figure 2. Multi-agent architecture for the Web services composition in the context of ebXML[27]

```
<process:hasPrecondition>
   <expr:KIF-Condition rdf:ID="ExistTGV">
      <expr:expressionBody rdf: datatype="http://www.w3.org/2001/XMLSchema#string">
            (ExistTGV ?From ?To)
      </expr:expressionBody>
   </expr:KIF-Condition>
</process:hasPrecondition>
```

The proposed architecture arranges of a set of sub-ontologies. Each sub-ontology includes all the concepts of certain type of Web services. For example, a sub-ontology$_i$ which represents the services of booking of any type of transport, another one which includes the Web services of booking of hotels, etc. We thought of this decomposition to improve the time answer to the requests.

### 5.1. ROLE OF ONTOLOGIES IN INTERACTION

As autonomous problem solvers, agents need to develop model of their environment that allows them to reason on how their actions affect their environment and how those changes lead them to achieve their goals [28]. Ontologies provide the conceptual framework that allows agents to construct such models: ontologies describe the properties of entities that agents encounter, and relations between them. Thus a common vocabulary in the form of ontologies is at the heart of intelligent communication among agents.

### 5.2. PRELIMINARY DEFINITIONS

In this section, we introduce some definitions necessary for the understanding of the proposed architecture functioning as well as that of the proposed composition model.

**Definition 5.1 (Operator)**

An operator can be defined by the quadruplet:
  o = (name(o), precond(o), add(o), del(o))
– name(o), the operator name, is defined by the following expression n(x1, . . . , xk) where n is Operator's symbol and x1, . . . , xk represent the parameters of the operator.
– precond(o) represent the preconditionc of the operator o, i.e., the world properties necessary for its execution.
– add(o) and del(o) define two sets of properties describing respectively the facts to be added and the facts to be deleted of the world state after the execution of o.

**Definition 5.2 (Action)**

An action is an instance of an operator. If $a$ is an action and $s_i$ is a state such as $precond^+(a)$ $s_i$ and $precond^-(a) \cap s_i = \emptyset$ then $a$ is applicable in $s_i$, and the result of this application is the state:
$$s_{i+1} = \gamma(s_i, a) = (s_i - \text{effets}^-(a)) \cup \text{effets}^+(a)$$

**Definition 5.3 (Planning domain)**

In planning, a domain defines all the operators who can apply to the world. A problem has to specify the initial state as well as the purpose to achieve.

A domain $D$ of $L$ is a restricted state transition system
$\Sigma = (S, A, \gamma)$ such as :
- $S = 2^{\{\text{instantiated atom of } L\}}$
- $A$ = {the set of instantiated operators of O} where O is the set of operators
- $\gamma(s, a) = (s - \text{effets}^-(a)) \cup \text{effets}^+(a)$ if $a \in A$ and $a$ is applicable in $s \in S$

**Definition 5.4 (Planning problem)**

A problem $P$ for a domain $D$ is a triplet P = $(O, s_0, g)$ where:
- $s_0$, the initial state, is some state of S;
- $g$, the goal, defines a coherent set of instantiated predicates, i.e., world properties must be reached;
- $O$ is the set of applicable operators.

**Definition 5.5 (Solution plan)**

A solution plan is defined as a linearization in a space of states. The passage from a state to the other one is made by the application of an action, i.e., an operator completely instantiated. Consequently, a solution plan for a planning problem $P = (O, s_0, g)$ is a sequence of actions describing a road of an initial state $s_0$ to a final state $s_n$. Such as the goal $g$ Is included in $s_n$. In other words, a plan $\pi$ is a solution for the problem $P$ if $\gamma(s_0, \pi)$ satisfy $g$.

**Definition 5.6 (Conjecture)**

A conjecture is a tuple $\chi = (A, <, I, C)$ such as:
- $A = \{a_0, \ldots, a_n\}$ is a set of actions.
- $<$ is a set of order constraints on the actions $A$ like $ai < aj$, i.e., $a_i$ precede $a_j$.
- $I$ is a set of instantiation constraints on variables of actions $A$ like x=y, x≠y or x=cst such as $cst \in D_x$ and $D_x$ is the domain of $x$.
- $C$ Is a set of causal links such $ai \xrightarrow{p} aj$ such as $a_i$ and $a_j$ are two actions of A, the order constraint $a_i < a_j$ exists in $<$, the property $p$ is an effect of $a_i$ and a precondition of $a_j$ and finally the instantiation constraints which connect variables of $a_i$ and of $a_j$ concerning the property $p$ are contained in $I$.

**Definition 5.7 (Hypothesis)**

Let's a conjecture $\chi=(A, <, I, C)$. A hypothesis formulate by $\chi$ is defined as a precondition $p$ of an action $a_j \in A$ such as for all actions $a_i \in A$, the causal link $ai \xrightarrow{p} aj \notin C$.

## 5.3. AGENTS ROLE AND TASKS

### 5.3.1. Manager agent

The manager agent makes sure that the description of an imported service is substantial with regard to the sub-ontology by controlling, in particular, its operations, its inputs and its outputs which have to be concepts of the corresponding sub-ontology. It stores all the OWL-S descriptions of services and the localizations of its suppliers. At the composition time, the manager agent corresponds to a Web service. It is initialized with the semantic description of the service and with a set of data allowing it reasoning. For example, an agent representing a service of booking of railroad transport has the list of the existing courses. These data form the knowledge base of the manager agent. From the semantic description, a manager agent is going to create its planning domain which includes, under the form of methods and operators extracted from the processes of the OWL-S description, the practicable actions by the agent, i.e., its base of competences.

### 5.3.2. Request Re-constructor Agent

The request re-constructor agent has for main task to reconstruct the "user request" from the descriptions recorded in the global ontology of the Web services. This one supports, at the same time, the atomic and composite services. The result of this task is an OWL-S file.

### 5.3.3. General Manager Agent

This agent has for main role to verify if the request represents the description of a composite service. In this case, the general manager agent invokes the composer agent, that invokes the agents concerned by the request. On the other hand, if the request user concerns an atomic service, the general manager agent invokes directly the concerned manager agent because it has a global view of the system.

### 5.3.4. Composer Agent

The composer agent has for role to coordinate and to assembly the manager agents to execute the required operations. It also initializes the composition process by sending to the manager agents the goal to be realized under the form of a first conjecture which represents the initial plan.

## 5.4. MAIN TASKS INVOLVED IN THE COMPOSITION PROCESS

The first task is user request construction. To make that, the user retrieves a special interface. After the user has finished its request construction, a request reconstruction step is made by the request constructor agent. To reconstruct a request, the request constructor agent retrieves OWL-S available services descriptions that are stored in the global BPs Ontology and which supports both atomic and composite processes.

Once the user request is semantically described, the general manager agent must be able to determine if the user request can be executed by one atomic service or by composing multiple services. In the second case, composer agent must be invoked. The composer agent processes the composite service following the order specified by the user; then it identifies the manager agents that manage the sub-ontologies that publish the required processes. It must be able to perform a matching between the inputs of successive operations and a matching between outputs of an invocation and the inputs of the next required one. In this case, we distinguish several types of compatibility [29] between the parameters. We have exact match and PlugIn match which represent a total matching, Subsume match and Fail match which give a partial matching. In our model, we consider total matching ("Exact match" and "PlugIn match"). We suppose the compatibility of IOPEs (Inputs, Outputs, Preconditions and Effects) at composition phase. Checking compatibility is made in selection services phase.

### 5.5. INTRODUCTIVE EXAMPLE

The following scenario allows illustrating what we mean by automatic and dynamic composition: a person X lives at Lyon and has to go to Tokyo for a conference. He decides to organize his travel by Internet by using two Web Services. Each service is represented by a manager agent: an Airways agent offering a service of plan tickets reservation and a Bank agent (representing X bank) that handles to pay the various reservations which X will be brought to realize. The problem that the user submits to the interface user can be summarized in the following way:
- Initial state: X is at Lyon
- Final state: X is at Tokyo

A possible plan resulting from the composition of these two Web Services can be expressed in an informal manner as follow:
1. Reserve the flight from Lyon to Paris.
2. Pay the ticket Lyon – Paris.
3. Reserve the flight from Paris to Tokyo.
4. Pay the ticket Paris - Tokyo.

Let us imagine now the dialogue that the various agents composing our architecture could build so that X can go to his conference:

**X** : « I'm at Lyon and I must be at Tokyo. Could you help me ? »
**General Manager Agent**: « I can't answer you immediately; I will ask that to the compositor agent. »
**Compositor agent:** « I will invoke the two manager agents Airways and Bank.»
**Airways Agent** : «I can take the user X to Tokyo provided that it is capable of going in Paris and I can take him from Paris towards Tokyo provided that he pays 150$ to go to Paris and 645$ to go to Tokyo»
**Bank Agent:** «Ok, I think that we keep the solution. I can pay the amount of 795$, the X account is credible. »

So the solution plan is: « Take the plan from Lyon to Paris then another one from Paris to Tokyo. » Its construction reposes on a centralized planning of the compositor agent with the cooperation of other planner agents (manager agents).

### 5.6. COMPOSITION MODEL

At the stage of the composition process [30], each manager agent represents a reserved service belonging to the sub-ontology that it manages. At the time of the composition, its objective is to simulate the service execution. This agent is an autonomous entity which contains a planner and which is capable to interact with the composer agent with the aim of co-building a plan of execution of Web services. The model which we propose is based on a multi-agent architecture whose manager agents represent the Web services. The manager agents are initialized with the semantic description of the service and with the knowledge base. The OWL-S description allows defining the planning domain of the agent whereas the knowledge base brings the knowledge which will be necessary for the agent to reason.

Our model leans on the dialectic plans synthesis proposed by [31] with centralization of decision-making (composer agent): the manager agents exchange the propositions, with the composer agent, in the form of sub-conjecture to build a solution plan. This model is constituted by three phases: the creation of the planning domain of every manager agent at the time of their initialization, the refinement of the proposed conjectures and sent to the composer agent, and the communication between composer agent and the manager agents, i.e., the submission of a new conjecture to the other agents further to the refinement of a conjecture.

### 5.6.1. Initialization of the manager agents and of the composer agent

Every manager agent corresponds to a reserved Web service. It is initialized with the semantic description of the service and with a set of data allowing the manager agent to reason. For example, an agent representing a service of booking of railroad transport has the list of the existing routes. These data form the knowledge base of the manager agent.

The composer agent is initialized with the semantic description of the user request got back with the general manager agent, and a set of data allowing it to reason.

### 5.6.2. Competences base Creation

From the semantic description, the manager agents create their planning domain which includes, under the form of methods and operators extracted from the processes of the OWL-S description, the practicable actions by the agents, i.e., their base of competences.

**a. Creation of the planning domain from the semantic description of a Web service**

The model of OWL-S process describes in a declarative way the properties and the behaviour of a Web service. The translation of the semantic description of a service towards a planning domain consists in representing the OWL-S processes in the form of operators and of methods. The algorithm of translation which we use is the one of [19]. The translation algorithm which we use is the one of [19]. This algorithm has as inputs an OWL-S description. The property "DescribedBy", which means that a service is described by a process, allows obtaining "the input point" from the description, i.e., the first process to be translated.

The algorithm of translation which we use is a recursive algorithm. Its principal is to transfer the first process in an operator if this last one is an atomic process Otherwise it translates recursively the participant processes in the control structures of this composite process and then creates the corresponding method.

**Example1.** For example, the following atomic process:

<process:AtomicProcess rdf:ID="AgentHotelReservation">
...
</process:AtomicProcess>
…………………………………….
Allow defining the following operator:
………………………………….. ;

```
(:operator (!AgentHotelReservation ...)
...
)
```

Then preconditions and effects of the atomic process are added to the corresponding operator.

**Example2.** This example demonstrates a method which corresponds to a composite process:

```
(:method (AgentFlightReservation ?AFR_From
?AFR_Date ?AFR_To ?AFR_CC)
( ... )
(
(!SearchFlight ?AFR_From ?AFR_To ?AFR_Date)
(!MakeReservation ?FlightID ?AFR_CC)))
```

This algorithm also solves the problems of management of the preconditions, the effects and the addition of goals to be reached.

### 5.6.3. Refinement of conjectures

A manager agent is going to argue, further to the conjecture received from the composer agent, from its competences (i.e., the actions which it is capable of planning) and of its knowledge to resolve the goal contained in this conjecture: it is going to refine it by adding to it a sub-conjecture, i.e., a sequence of actions, or by adding causal links. A manager agent can refine a conjecture by sending hypotheses on the properties which he does not know. These hypotheses form new goals to resolve for the other manager agents.

### 5.7. AGENTS ARCHITECTURE AND FUNCTIONNING

### 5.7.1. Request constructor agent

In the early stage of the Web, information was shared as HTML pages. These pages were designated to be read only by a human user [32]. The first language designed by the consortium W3C in the domain of Web Semantic is the RDF (Resource Description framework) language [33]. RDF is an XML language used for describing metadata and for facilitating their treatment by specific applications programs. RDFS (RDF Schema) language was developed after in order to give RDF more expressive power. However, many limitations restrict the ability to express knowledge. Indeed, it is not possible to carry out an automated reasoning on knowledge modelled using RDFS. To overcome this lack, a new language for Web called OWL was developed (Ontology Web Language) [34]. OWL is based on logic description. Using OWL, one can describe the knowledge about a domain in terms of a set of classes and a set of properties. Classes represent entities of interest in a specific domain and a property represents a feature (i.e. data type property) of an entity or a relationship between entities (i.e. object properties). OWL, like RDF, is based on XML language. OWL provides tools for comparing and reasoning on classes, their features and the relations between them. It gives a great ability to interpret the web content because it contains a wide range of vocabulary and a full semantic formal. The W3C provided three types of languages to better express OWL: *Lite*, *DL* (Description Logic) and *Full*.

The objective of this agent is to produce an OWL-S file from a simple user request which is represented through an XML document. This production represents the description of an instance of a specified service that corresponds to the customer request. It represents the input of the Request Constructor Agent (cf. Figure 3).

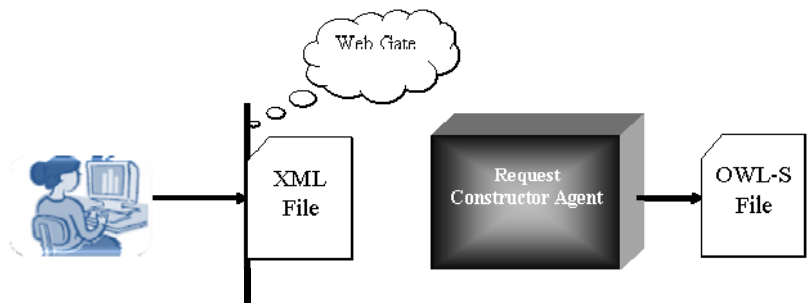

Figure 3. Global View of the Request Constructor Agent

The architecture of this agent is shown in Figure 4:

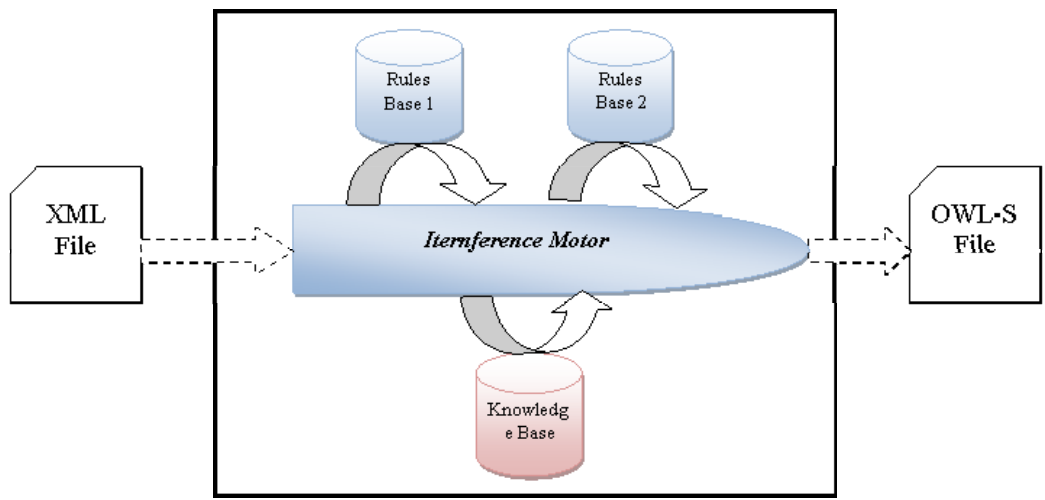

Figure 4. Request Constructor Agent Architecture

The input of the agent is an XML file which represents the customer request. When a user sends his request by filling the service formulary interface, this request will be directly represented under the XML standard with the values which the user has introduced at first time, what is going to represent the input file. This file will be received by the Request Constructor Request.

**A. Request Constructor Agent Functioning**

It is all the treatment which the agent makes to produce the result which is the description of an instance of a web service, this treatment consists of four phases: *Crossing the XML file*, *Generation of instances*, *Extraction of a part of agent sub-ontology*, and the *web service description production*. Thus we can say that all the spots which follow each other compose the inference engine of the agent (cf. Figure 5).

**B. Tasks Involved in the Construction Request Process**

The function of the Constructor Request Agent is divided in four main tasks (cf. Figure 5): *crossing the XML file*, *generation of instances*, *extraction of specific ontology* and *production of the service description*.

**B1. Crossing the XML file and generation of instances**

The Request Constructor Agent crosses the XML file to extract its arborescence (cf. Figure 6). Then it sends these last ones to the rules base N°1, in order to know those representing classes and those represent attributes. So, the rule that uses the agent in this case is as follow: "If a tag possesses sub-tags then it is considered as a class, otherwise it is an attribute. Then by applying this rule, we obtain the names of classes with its attributes and also their values (instances)".

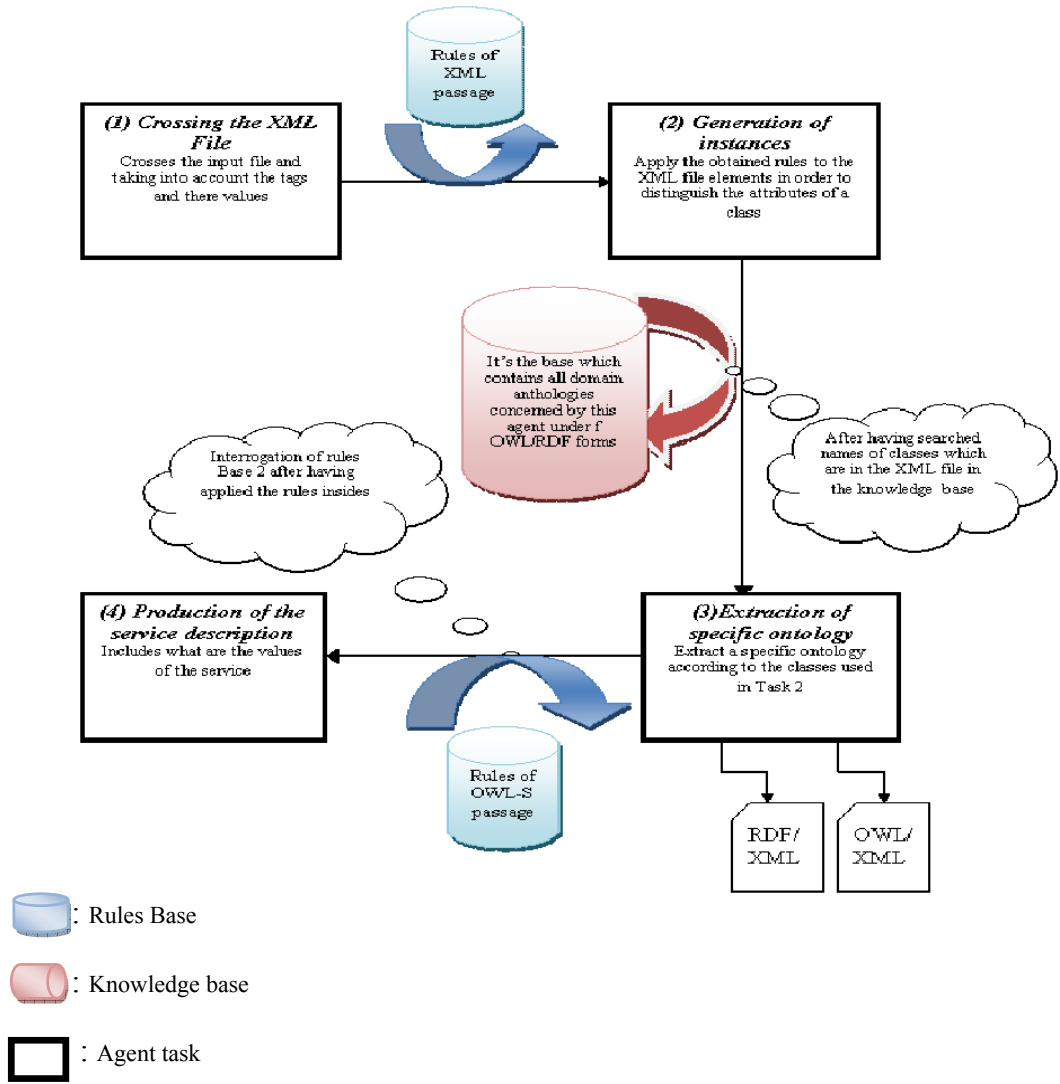

Figure 5. Request Constructor Agent Process

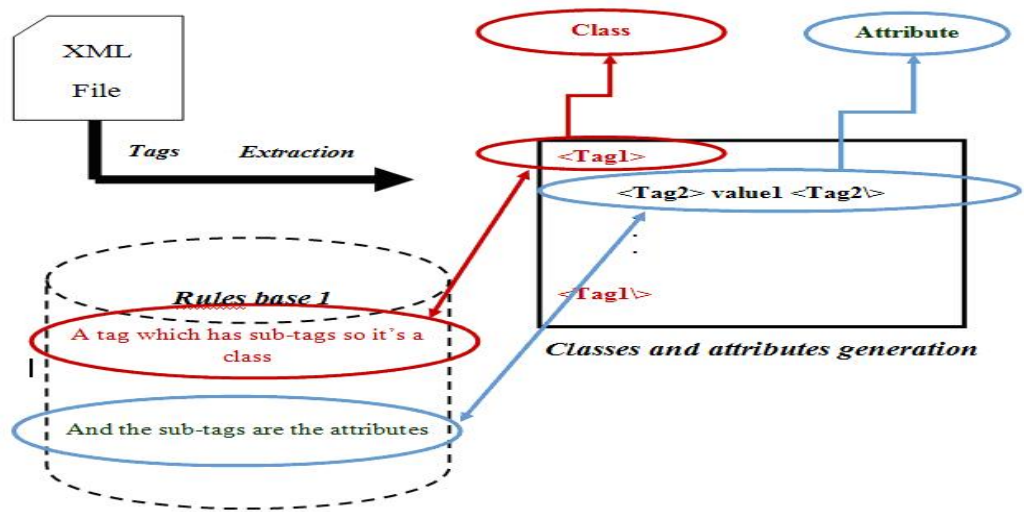

Figure 6. Crossing the XML file and instances generation

**B2. Extraction of a specific ontology**

The Request Constructor Agent has a knowledge base which represents all ontologies that exist as well as ontologies which the Agent knows during his experience. The knowledge base gives us classes obtained from the XML file in order to know the relations and classes which have direct relations with the classes produced by the previous phase, so we obtain a file which contains a more complete structure of classes. New relations and classes are added to enrich the knowledge base of the agent.

So in this phase, the agent extracts a specific ontology according to the classes used in Tasks 2 (File OWL / XML). In order to produce a more readable and convivial OWL / XML file, we have seen that it is recommended to make a small transformation of OWL / XML to pass in the RDF / XML format. For that, we have defined two main rules (cf. Figure 7 and Figure 8) (rules base N°2):

***Rule 1:*** "*If we find classes in the XML file obtained in the entry and if they are scattered in various ontologies, then the produced service considered as the output of the agent will be a composed service*".

In a different way, a simple service is a service of which all the classes belong to the same ontology.

***Rule 2:*** "*The attributes values of the obtained instances are the inputs of the service, and the empty attributes (without values) represent its outputs*".

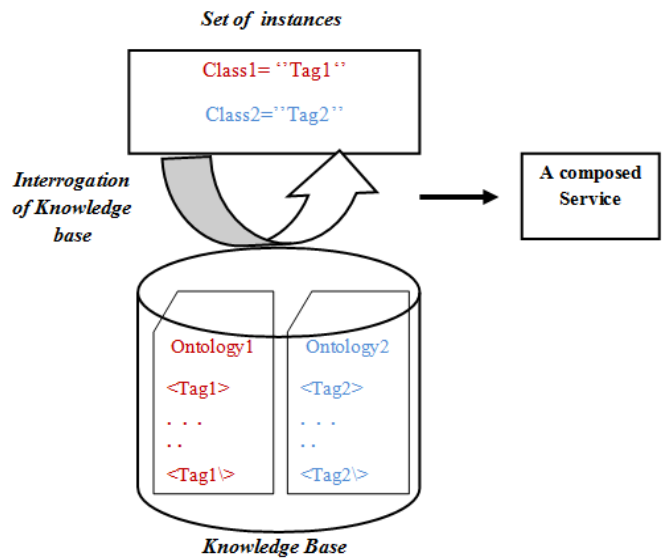

Figure 7. Rule 1 of rules base N°2

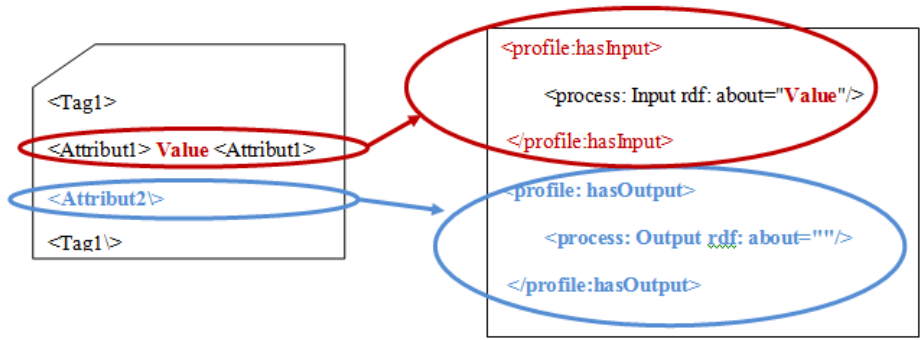

Figure 8. Rule 2 of rules base 2

**B3. Description Service Production**

After having applied the rules of the Rules base N°2, we can decide and understand what will be the service type i.e. we can produce the service description corresponding to the user request(by affecting the XML file values). So, the result is an OWL-S file to send to the General Manager Agent.

**5.7.2. General Manager Agent**

The General Manager Agent is capable of reasoning about the agent that may be asked for achieving the goal (the asked agent can be a manager one or the composer one according to the web service type indicated in the user request) and must be capable of communicating with the other agents (manager's one or composer one). It is initialized with the semantic description of the user goal to be realized (an OWL-S file received beside the Request constructor agent).

The General Manager Agent (cf. figure 9) consists of two main modules:
- The reasoning module.
- The dialogue administrator who allows the dialogue with the other agents.

**A. Reasoning module**

The reasoning module is the main module of the agent. It defines the behaviour of the agent by managing its interactions. It leans on the administrator of dialogue to submit the request to the required agent.

The General Manager Agent will, at the first time, cross the OWL-S file then it will detect if the required service is a composed one or an atomic one.

**A1. The required service is an atomic one.** If he does not find the tag that indicates that the service is composed *<process :CompositeProcess>*, then he will deduct that the service is an atomic one (in this case he will find *<process: AtomicProcess ....>*).

**Example of an atomic service.**

```
<process:AtomicProcess rdf:ID="LogIn">
   <process:hasInput rdf:resource="#AcctName_In"/>
   <process:hasInput rdf:resource="#Password_In"/>
</process:AtomicProcess>
<process:Input rdf:ID="AcctName_In">
   <process:parameterType rdf:resource="&concepts;#AcctName">
</process:Input>
<process:Input rdf:ID="Password_In">
   <process:parameterType rdf:resource="&concepts;#Password">
</process:Input>
```

In this case, the General Manager Agent will select the manager agent concerned by the request; this later is made from a selection table which the structure is the following one:

Table 1. Structure of the selection table

| Agents Group | Supplied service | Available agents sorted out according to the performances |
|---|---|---|
| 1 | Air transport | a1, a12, a24 |
| 2 | Reservation of hotels | a34 |
| 3 | Payment service | a101, a3, a2 |

After selection of the adequate manager agent, the General Manager Agent sends it a message (the OWL-S file) indicating the asked service with the parameters fixed by the user.

**A2. The required service is a composed one.** In this case, the General Manager Agent finds the tag *<process:CompositeProcess>* which indicates that the user request requires the composition of two or several web services indicated between the two tags *<process:composedOf>* and *</process:composedOf>*. Thus, the General Manager Agent will deduct the manager agents implied in the composition process because he has a global view of the system. So, it is going to question the composer agent by sending it the group of manager agents concerned by the request as well as the list of the manager agents available at present (we do not manage the dynamic availability and not availability of the various manager agents because it's a very complex problem).

**Example of a composed service.**

```
<process:CompositeProcess rdf:ID="BravoAir_Process">
 <rdfs:label>This is the top level process for BravoAir</rdfs:label>
```

```xml
  <process:composedOf>
    <process:Sequence>
      <process:components rdf:parseType="Collection">
        <process:AtomicProcess rdf:about="#GetDesiredFlightDetails"/>
        <process:AtomicProcess rdf:about="#SelectAvailableFlight"/>
        <process:CompositeProcess rdf:about="#BookFlight"/>
      </process:components>
    </process:Sequence>
  </process:composedOf>
</process:CompositeProcess>

<process:CompositeProcess rdf:ID="BookFlight">
 <process:composedOf>
  <process:Sequence>
   <process:components rdf:parseType="Collection">
    <process:AtomicProcess rdf:about="#Login"/>
    <process:AtomicProcess rdf:about="#ConfirmReservation"/>
   </process:components>
  </process:Sequence>
 </process:composedOf>
</process:CompositeProcess>
```

**B. Dialogue administrator module**

The dialogue administrator is the module which allows exchanging messages between the composer or the manager agents.

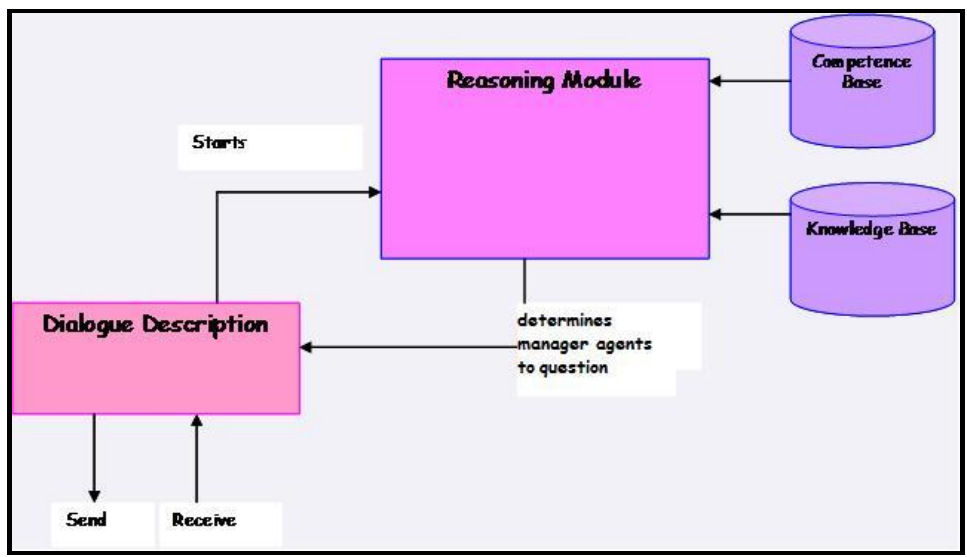

Figure 9. General Manager Agent Architecture

### 5.7.3. Composer agent

The composer agent must be capable of reasoning about the task that he has to achieve and must be capable of communicating with the other agents (manager's one). It is initialized with the semantic description of the user goal to be realized and the group of manager agents sent beside the General Manager Agent.

The composer agent (cf. figure 10) consists of t²hree main modules [30]:

- The reasoning module.
- The storage board which serves as a support of the conjectures received from the manager agents.
- The dialogue administrator who allows the dialogue with the other agents.

**A. Storage board**

The storage board is used as a support to the dialogue by saving the sub-conjectures proposed by the various manager agents.

**B. Dialogue administrator**

The dialogue administrator is the module which allows exchanging the propositions between the composer agent and the manager agents.

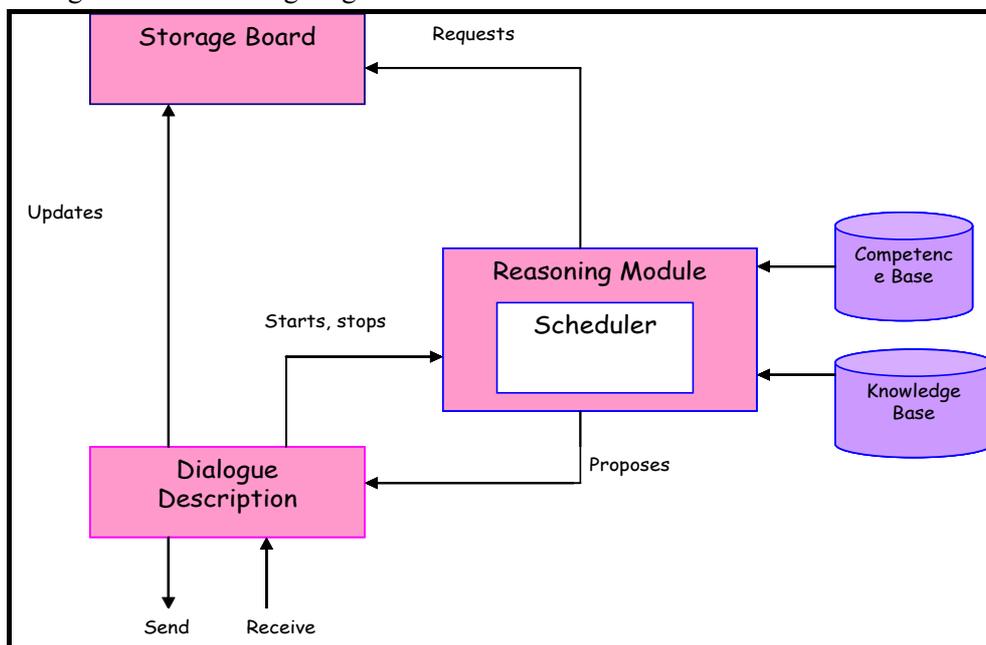

Figure 10. Composer Agent Architecture [30]

**C. Reasoning module**

The reasoning module is the main module of the agent. It defines its behaviour by managing its interactions. It leans on i) the storage board to select the conjecture to be refined and to select the subset of the manager agents with whom it will continue the process of refinement and on ii) the administrator of dialogue to submit the propositions to the agents.

The composition dialogue is initiated by the composer agent who subjects to all concerned manager agents a first conjecture. This conjecture constitutes the user goal to be realized. At the reception of this first conjecture, the manager agents begin to refine it.

When a manager agent calculates a sub-conjecture, it submits it to the composer agent. At the reception of a refinement, the composer agent updates his storage board. Then he tries to send back the new sub-conjecture to refine to the manager agents whom he holds for the next cycle of refinement.

### 5.7.4. Manager agent

A manager agent (cf. Figure 11)[30] consists of two main modules:
- The reasoning module which contains a planner.
- The dialogue administrator module.

#### A. Reasoning module

The main behaviour of a manager agent is the refinement of the conjectures received from the composer agent and the sending, to this last one, its propositions (refinements). The refinement of a conjecture can be made of two manners:
- By adding of causal links. First, the manager agent verifies that there is, in the conjecture, no action allowing realizing the goal. If that was the case, then a causal link is added between the effect concerned by the first action and the precondition of the second action.
- By adding of a sub-conjecture. The manager agent tries to resolve the hypothesis by producing a sub-conjecture by means of the planner. The planner takes as inputs the initial state, i.e., the knowledge base of the agent, as well as the goal to be resolved and supplies, if it exists, a solution plan for this goal.

The obtained plan, enriched by constraints of orders and causal links, forms the solution sub-conjecture of the hypothesis and will be sent to the composer agent.

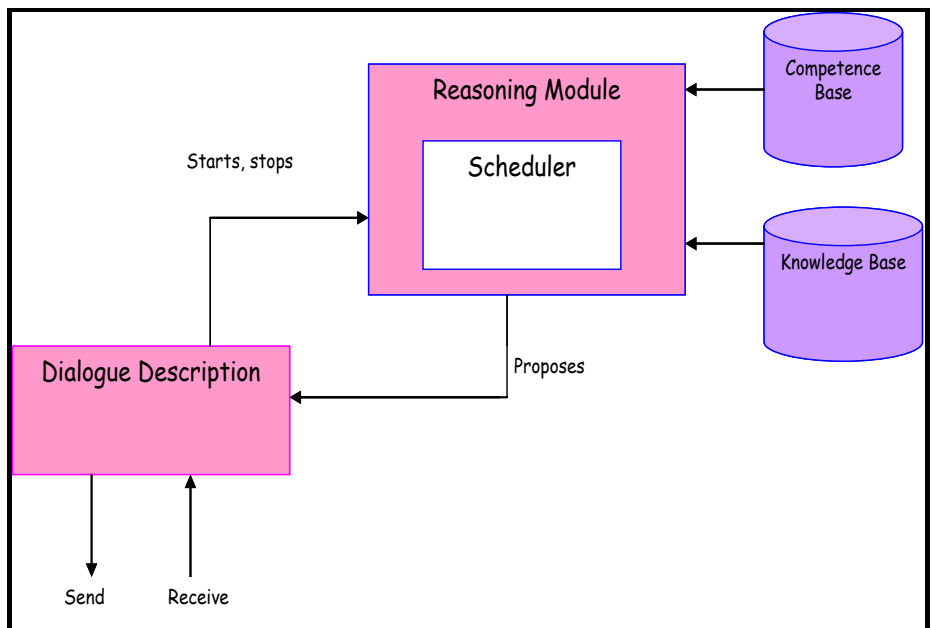

Figure 11. Manager Agent Architecture [30]

#### B. Dialogue administrator module

It is the module which allows exchanging the propositions between the manager and the composer agents.

# 6. CONCLUSION

The Web service composition problem is a challenging research issue because of the growth in the number of Web services available, the dynamic environment and changing user needs.

We have presented an agent based architecture that, dynamically, composes Semantic Web Services. This architecture allows defining different levels of responsibility:
-constructing semantically the user request assured by the request constructor agent according to the global BPs ontology,
- Attribution a task to composer agent or directly to manager agent assured by the general manager agent according to its global view of the system,
- composing and coordinating the composite service execution assured by the composer agent.

The proposed composition model is based on the distributed planning with the dialectical synthesis of plans under hypotheses. The originality of our model is that the solution plan succeeding is not totally distributed because of the use of a composer agent who collaborates with various manager agents (which present Web services used in the composition process). The idea is that there is always an agent who oversees the progress state of the composition process and who has a global vision allowing to localize the problem in a failure case. We have implemented the *Request Constructor Agent* by using several packages such as: JDOM, OWL-API, OWL-S_API. The proposed architecture can be implemented within the ebXML functional specification for ebXML BPs composition.

Future work will focus mainly on managing the dynamic availability and not availability of the various manager agents.

## REFERENCES

[1] A.Ben Hassine, Shigeo Matsubara & Toru Ishida, (2006) "A Constraint-based Approach to Horizontal Web Service Composition", 5$^{th}$ International Semantic Web Conference (ISWC2006), Athens, GA, USA, November 5-9, 2006.

[2] OWL Services Coalition, (2003)" OWL-S: Semantic markup for web services", OWL-S White Paper http://www.daml.org/services/owl-s/1.0/owl-s.pdf.

[3] A. Ankolekar., M.Burstein., J.R.Hobbs., O.Lassila., D.L.Martin, D.McDermott, S.A. McIlraith., S.Narayanan, M.Paolucci, T.R.Payne & K.Sycara, (2002) "DAML-S: Web Service Description for the Semantic Web", In Horrocks, I., Hendler, J., eds.: Proceedings of the First International Semantic Web Conference (ISWC2002), Sardinia, Italy. Volume 2342 / 2002 of Lecture Notes in Computer Science., Springer-Verlag Heidelberg (2002),pp 348-363.

[4] K.Sycara, M.Paolucci, A. Ankolekar & N. Srinivasan (2003) "Automated discovery, interaction and composition of semantic web services", Journal of Web Semantics **1** (2003), pp 27-46

[5] D.Wu, E.Sirin, B.Parsia, J.Hendler & D.Nau (2003) "Automatic web services composition usingSHOP2", In: Proceedings of ICAPS'03Workshop on Planning for Web Services, Trento, Italy.

[6] M.Pasha and H. Farouq Ahmad (2008) "Agents Negotiating with Semantic Web Services", Proceedings of the World Congress on Engineering and Computer Science 2008WCECS 2008, October 22 - 24, 2008, San Francisco, USA.

[7] Web Services Architectures, W3C Working Draft 14 May 2003: http://www.w3.org/TR/2003/WD-ws-arch-20030514/

[8] M.Lyell, L.Rosen, L.Casagni-Simkins & D.Norris (July, 2003) "On software agents and web services", In Proc. of the 1st International Workshop on Web Services and Agent Based Engineering, Sydney, Australia.

[9] E.M. Maximilien, M.P. Singh, M., P. (July, 2003) "Agent-based architecture for autonomic web service selection", In Proc. of the 1st International Workshop on Web Services and Agent Based Engineering, Sydney, Australia.

[10] T.Berners-Lee, J.Hendler & O.Lassila (2001) "The semantic web. Scientific American" 284(5): pp 34–43.

[11] W3C Semantic Web Activity: http://www.w3.org/2001/sw/.

[12] S. McIlraith. & T.C.Son (2002) " Adapting Golog for Composition of Semantic Web Services", KR- 2002, France, 2002.

[13] E.Sirin, B.Parsia, D.Wu, J.Hendler & DNau (2004) " HTN Planning for Web Service Composition Using SHOP2". In Journal of Web Semantic Vol. 1, pp. 377-396.

[14] M. Lin, J.Xie, H.Guo & H.Wang,(2005) "Solving Qos-driven Web Service Dynamic Composition as Fuzzy Constraint Satisfaction", *In proc. IEEE Int. Conf. on e-Technology, e-Commerce and e-service, EEE'05*, pp. 9-14.

[15] T.Ishida (2006) "Language Grid: An Infrastructure for Intercultural Collaboration. Valued Constraint Satisfaction Problems: Hard and Easy Problems", In IEEE/IPSJ Symposium on Applications and the Internet (SAINT-06), pp. 96-100.

[16] M. Pistore, F. Barbon, P. Bertoli, D. Shaparau & P. Traverso (2004) "Planning and MonitoringWeb Service Composition", Proceedings of the International Conference on Artificial Intelligence, Methodology,Systems, and Applications, p. 106-115.

[17] M. Pistore, A. Marconi, P. Bertoli & P. Traverso (2005) "Automated Composition of Web Services by Planning at the Knowledge Level", Proceedings of the International Joint Conference on Artificial Intelligence, p. 1252-1259.

[18] M. Pistore, P. Traverso & P. Bertoli (2005) "Automated Composition of Web Services by Planning in Asynchronous Domains", Proceedings of the International Conference on Planning and Scheduling, pp. 2-11.

[19] J.Guitton (2006) "Planification multi-agent pour la composition dynamique de services web ", Master project, Joseph Fourier University, France.

[20] D. Pellier &H. Fiorino (2009) "Un modèle de composition automatique et distribué de services Web par planification", RSTI - RIA - 23/2009. Intelligence artificielle et web intelligence, pp 13-46.

[21] S. McIlraith, S.Son & H. Zeng (2001) "Semantic Web Services", IEEE Intelligent Systems, 16(2): pp46– 53.

[22] M. Sheshagiri, M. desJardins & T. Finin (2003) "A Planner for Composing Services Described in DAML-S", In Proc. Of Workshop on Web Services and Agent-based Engineering - AAMAS'03.

[23] M. Jayadev & C. William (2007) "Computation Orchestration: A Basis for Wide-area Computing", Journal of Software and Systems Modeling, vol. 6, n° 1, pp 83- 110.

[24] E.Sirin J. Hendler &B. Parsia (2003) "Semi-automatic composition of web services using semantic descriptions", In Proceedings Web Services: Modeling, Architecture and Infrastructure. Workshop in Conjunction with ICEIS2003 (Angers, France, 2002), ICEIS Press, pp17-24.

[25] B. Medjahed, A. Bouguettaya & A. Elmagarmid (November 2003) "Semantic web enabled composition of web services", The VLDB Journal 12, 4, pp333–351.

[26] ebXML (2002), http://www.ebXML.org/specs/ebTA.pdf.

[27] O. Hioual & Z. Boufaida (2008) "Towards a Semantic Composition of ebXML Business Processes"*,* International Conference on Innovations in Information Technology, 2008. IIT 2008, Dubai, 16-18 Dec. 2008, pp165-169.

[28] K. Sycara & M.Paolucci (2001) "Ontologies in Agent Infrastructure". Carnegie Mellon University, USA, 2001.

[29] A. El-Fallah-Seghrouchni & S. Haddad. (1996) "A coordination algorithm for multi-agent planning", In Lecture Notes in Computer Science, vol. 1038. Springer Verlag Publisher, 1996, pp86–99.

[30] O. Hioual & Z. Boufaida (2010) "Vers une architecture à base d'agents pour une composition sémantique et dynamique des services web dans un contexte d'ebXML", 8th ENIM/IFAC International Conference on Modelling and Simulation - MOSIM'10 – May, 10-12, 2010 - Hammamet – Tunisia.

[31] D.Pellier (2005) "Modèle dialectique pour la synthèse de plans", PhD thesis, UJF - Grenoble, France, 2005.

[32] G. Nachouki & M.P. Chastang (2010) "Multi-data source fusion approach in Peer-to-Peer systems", The International Journal of Database Management Systems (IJDMS), Vol.2, No.1, February 2010.

[33] RDF: Resource Description Framework http://www.w3.org/RDF/

[34] D.L. McGuinness, F. Van Harmelen, "OWL Web Ontology Language Overview", in: W3C Recommendation, W3C, 2004